\documentclass[10pt,twocolumn,letterpaper]{article}

\usepackage{cvpr}              
\definecolor{cvprblue}{rgb}{0.21,0.49,0.74}
\usepackage[pagebackref,breaklinks,colorlinks,allcolors=cvprblue]{hyperref}
\usepackage[utf8]{inputenc}
\def\paperID{9277} 
\def\confName{CVPR}
\def\confYear{2026}
\usepackage{multirow}
\usepackage[table]{xcolor} 
\newcommand{\best}[1]{\textbf{#1}}
\newcommand{\second}[1]{#1}

\title{Beyond Darkness: Thermal-Supervised 3D Gaussian Splatting for Low-Light Novel View Synthesis}

\author{
Qingsen Ma\thanks{These authors contributed equally.} \quad
Chen Zou\footnotemark[1] \quad
Dianyun Wang\footnotemark[1] \quad
Jia Wang\footnotemark[1] \quad
Liuyu Xiang \quad
Zhaofeng He\thanks{Corresponding author.}\\
Beijing University of Posts and Telecommunications\\
{\tt\small maqingsen@bupt.edu.cn}
}

\begin{document}
\maketitle
\begin{abstract}
Under extremely low-light conditions, novel view synthesis (NVS) faces severe degradation in terms of geometry, color consistency, and radiometric stability. Standard 3D Gaussian Splatting (3DGS) pipelines fail when applied directly to underexposed inputs, as independent enhancement across views causes illumination inconsistencies and geometric distortion. To address this, we present DTGS, a unified framework that tightly couples Retinex-inspired illumination decomposition with thermal-guided 3D Gaussian Splatting for illumination-invariant reconstruction. Unlike prior approaches that treat enhancement as a pre-processing step, DTGS performs joint optimization across enhancement, geometry, and thermal supervision through a cyclic enhancement–reconstruction mechanism. A thermal supervisory branch stabilizes both color restoration and geometry learning by dynamically balancing enhancement, structural, and thermal losses. Moreover, a Retinex-based decomposition module embedded within the 3DGS loop provides physically interpretable reflectance–illumination separation, ensuring consistent color and texture across viewpoints. To evaluate our method, we construct RGBT-LOW, a new multi-view low-light thermal dataset capturing severe illumination degradation. Extensive experiments show that DTGS significantly outperforms existing low-light enhancement and 3D reconstruction baselines, achieving superior radiometric consistency, geometric fidelity, and color stability under extreme illumination.
\end{abstract}

\section{Introduction}
\begin{figure}
    \centering
    \includegraphics[width=1\linewidth]{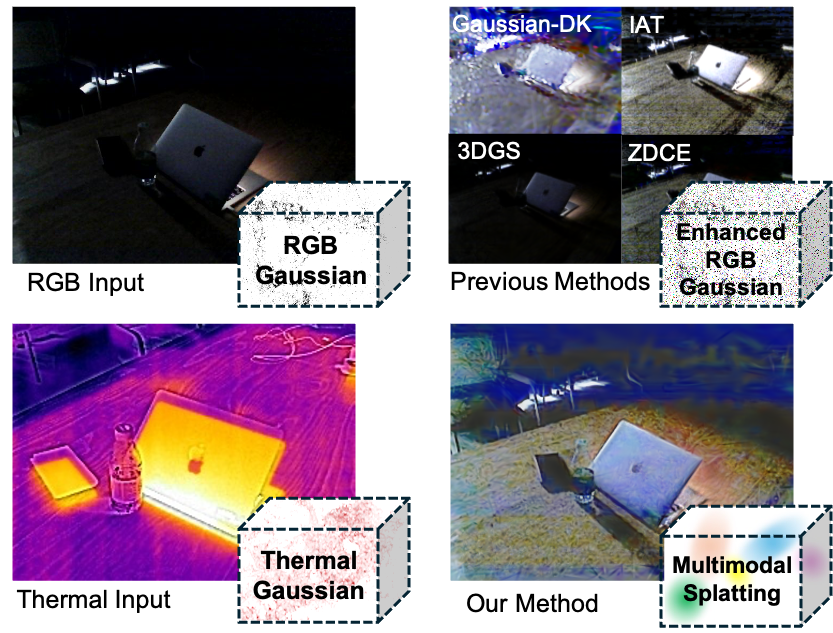}
    \caption{RGB-only enhancement struggles to maintain structural coherence and color consistency in extremely low-light scenes. Our method delivers more consistent colors and superior contrast.}
    \label{fig:placeholder}
\end{figure}

Novel View Synthesis (NVS) is a fundamental task in computer vision with broad applications in augmented and virtual reality (AR/VR)~\cite{li2207compnvs,habtegebrial2018fast,connor2002novel,elata2025novel}. Recent advances such as Neural Radiance Fields (NeRF)~\cite{mildenhall2021nerf} and 3D Gaussian Splatting (3DGS)~\cite{kerbl20233d} have achieved high-quality rendering and impressive efficiency, enabling real-time or near real-time performance. However, these successes typically rely on multi-view images that are well-exposed and of high signal-to-noise ratio. In real-world scenarios—e.g., archaeological exploration in dark caves, nighttime military simulation, nocturnal driving, or robotic search-and-rescue in poorly lit environments—illumination is harsh or extremely limited, and existing NVS pipelines~\cite{mildenhall2022nerf, wang2023lighting} struggle to preserve color consistency, structural fidelity, and geometric interpretability. 

Directly applying 3DGS to images captured under adverse illumination is challenging for several reasons. First, severe under-exposure, significant noise, and color distortion cause substantial information loss, making it difficult to recover high-fidelity geometry and radiometry in the absence of ground-truth references. Second, naively combining exposure-correction or low-light enhancement~\cite{guo2020zero, li2021learning, zhang2021low} with 3DGS often introduces cross-view illumination inconsistencies: each view is enhanced independently, so inter-view lighting statistics diverge, degrading multi-view consistency and reconstruction quality. 
Third, extreme darkness exacerbates color shifts and structural blurring; even state-of-the-art enhancement models ~\cite{cai2023retinexformer,tang2022improved,yang2018low}can produce color artifacts, hue drift, or over-smoothing that harm downstream 3D reconstruction. 

To address these challenges, we introduce a modality that is inherently invariant to illumination and provides a consistent supervisory signal during training. Thermal imaging naturally exhibits this property. Unlike RGB cameras that capture reflected light, thermal sensors measure emitted radiation, offering reliable signals under low-light or even no-light conditions. Consequently, thermal data preserves stable contours and physically grounded spatial distributions correlated with material and temperature variations. We hypothesize that leveraging this illumination-invariant modality as a stable anchor can jointly guide RGB enhancement and 3D reconstruction toward a unified and consistent solution~\cite{hassan2024thermonerf}.

Based on this insight,We propose \textbf{DTGS}, a novel framework that tightly couples Retinex-inspired illumination enhancement~\cite{land1971lightness} with end-to-end thermal-guided 3D Gaussian Splatting. 
Unlike prior pipelines that treat Retinex-style enhancement as a fixed pre-processing step, DTGS adopts a \emph{joint} optimization strategy: illumination enhancement, 3D reconstruction, and thermal-specific consistency are optimized together. By closely coupling these components, DTGS can recover both structural and radiometric information from severely under-exposed inputs and maintain inter-view consistency even under the most challenging conditions.

The core innovation of DTGS lies in its integration of a \emph{thermal supervisory branch} that jointly guides and stabilizes both Retinex-based enhancement and 3D Gaussian Splatting (3DGS) reconstruction. 
Within the cyclic enhancement–reconstruction process, the thermal modality provides illumination-invariant structural cues that constrain both reflectance recovery and geometric optimization. 
Meanwhile, an \emph{adaptive weight scheduling} strategy dynamically balances the contributions of enhancement, geometric, and thermal losses, ensuring that illumination correction and 3D modeling evolve cooperatively rather than competitively. 
This unified design enables DTGS to achieve stable color restoration, robust radiometric consistency, and accurate geometry under extreme low-light conditions.

To rigorously evaluate DTGS, we construct a multi-view low-light thermal 3D reconstruction dataset. 
While existing datasets have explored NVS from raw noisy images~\cite{lu2024thermalgaussian,cui2024alethnerfilluminationadaptivenerf}, a dedicated benchmark for thermal-guided 3D reconstruction under such adverse conditions is still lacking.
Our dataset spans extremely adverse degradations, including ultra-low dynamic range, severe noise, and cross-view radiometric inconsistency, providing a realistic and challenging benchmark. Qualitative comparisons show that DTGS surpasses representative enhancement-plus-reconstruction baselines, achieving superior geometric fidelity and radiometric stability. These findings suggest that DTGS sets a new bar for robust 3D reconstruction in extremely low-light environments.

Our contributions are as follows:
\begin{itemize}

\item We propose a framework that tightly couples 3DGS with Retinex-based enhancement, trained under the joint supervision of a thermal modality, tailored for extreme low-light environments.

\item We embed illumination-reflectance decomposition directly into the 3DGS loop, enabling physically-interpretable enhancement that co-evolves with geometry under thermal supervision.

\item We propose a dynamic ground-truth update strategy that effectively breaks the enhancement-geometry deadlock by allowing the 3D model to train on progressively cleaner, internally-generated targets.

\item We construct and release a challenging multi-view dataset, providing a realistic benchmark for evaluating illumination-aware and thermally-guided 3D reconstruction under adverse lighting conditions.

\end{itemize}

\section{Related Work}

\subsection{Low-Light Enhancement.}
Recent extensions of 3D Gaussian Splatting (3DGS), such as LITA-GS~\cite{zhou2025lita} and Luminance-GS~\cite{cui2025luminance}, improve cross-view consistency via illumination-robust priors and view-adaptive tone curves, yet remain limited under extreme low light where SNR collapses and radiometric cues degrade. Gaussian-DK~\cite{ye2024gaussian} incorporates exposure-compensated supervision for dark scenes but still suffers from color and geometric instability, highlighting the difficulty of jointly preserving photometry and structure. Retinex theory~\cite{land1971lightness}, including MSR/MSRCR, and its deep or curve-based variants such as RetinexNet~\cite{wei2018deep}, Zero-DCE~\cite{guo2020zero}, and ISP/Transformer approaches~\cite{wang2023ultra, cui2022you}, decompose images into reflectance and illumination. This decomposition is well suited for multi-view 3DGS: it separates exposure-variant illumination from exposure-invariant reflectance, aligns naturally with Gaussian color/opacity optimization, and allows priors like smooth illumination and sparse reflectance—together with auxiliary multi-view constraints (e.g., thermal histograms/gradients)—to prevent hallucination and maintain consistency in extreme low-light conditions.

\subsection{Thermal Modality and Its Role in 3D Modeling}
Thermal imaging complements RGB for 3D reconstruction by capturing emitted radiation, making it illumination-invariant and robust in low/no light~\cite{vidas20133d, weinmann2014thermal}. Early work adapted SfM/SLAM to thermal inputs for point clouds and surfaces~\cite{yamaguchi2017application, de20223d}, demonstrating geometric feasibility but limited texture fidelity due to lower resolution and higher noise.

Neural scene representations deepen these advances. NeRF variants fuse RGB/thermal streams or use thermal cues to guide geometry for robust novel-view synthesis~\cite{xu2024leveraging, hassan2024thermonerf}. Gaussian splatting has likewise been extended with thermal data for real-time, high-quality rendering~\cite{chen2024thermal3d}. Cross-spectral formulations jointly model radiance and geometry with automatic pose calibration~\cite{poggi2022cross}; Thermal Gaussian~\cite{chen2024thermal3d}, HyperGS~\cite{thirgood2025hypergs}, and SpectralGaussians~\cite{sinha2024spectralgaussians} further improve radiometric consistency and geometric fidelity. Building on these, our method treats thermal signals as supervision within a 3DGS framework: thermal edges anchor Gaussian centers/covariances, histogram–gradient statistics enable per-view radiometric calibration, and high-SNR thermal gradients drive adaptive loss reweighting—stabilizing Retinex-style enhancement and reconstruction to yield color-consistent, geometrically faithful results under extreme low light.

\section{Method}

We propose DTGS, a 3D reconstruction framework tailored for low-light thermal imaging. Instead of relying on pixel-level enhancement, DTGS employs a Retinex-based decomposition that aligns with 3D Gaussian Splatting for joint optimization of reflectance, illumination, and geometry. To ensure consistency, we introduce a thermal consistency loss and a cyclic ground-truth refresh strategy, mitigating the “enhancement–geometry deadlock” and improving both radiometric and geometric fidelity. Thermal imaging further provides illumination-invariant structural cues, stabilizing the enhancement–reconstruction process and enabling sharper, more consistent results in extreme low-light scenes.

\begin{figure*}
    \centering
    \includegraphics[width=1\linewidth]{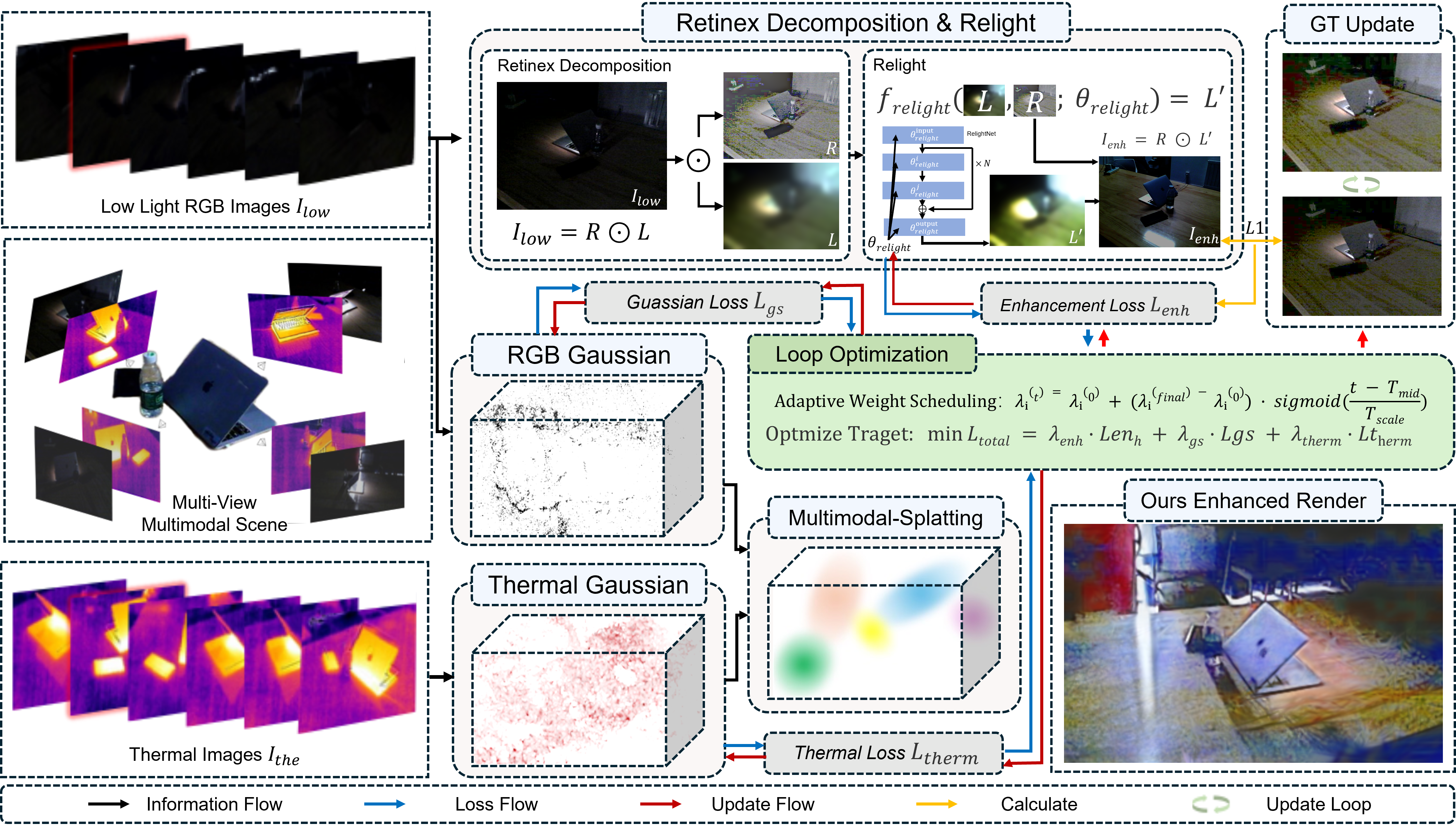}
    \caption{Overview of the DTGS Framework.
    Given multi-view low-light and thermal images of a scene, our method performs 
    joint enhancement and 3D reconstruction through cyclic optimization. 
    The thermal modality guides Retinex-based enhancement of low-light images, 
    producing $I_{\text{enh}}$ that progressively updates the ground truth 
    $GT^{(t)} = (1-\alpha^{(t)}) \cdot GT^{(t-1)} + \alpha^{(t)} \cdot I_{\text{enh}}^{(t)}$ 
    for 3D Gaussian Splatting training. 
    The total loss 
    $\mathcal{L}_{\text{total}} = \lambda_{\text{enh}} \cdot \mathcal{L}_{\text{enh}} + \lambda_{\text{gs}} \cdot \mathcal{L}_{\text{gs}} + \lambda_{\text{therm}} \cdot \mathcal{L}_{\text{therm}}$ 
    is optimized with adaptive four-stage weight scheduling. 
    Thermal consistency loss ensures cross-modal coherence, enabling bidirectional 
    gradient flow between enhancement and reconstruction modules.}
    \label{fig:pipeline}
    \label{fig:placeholder}
\end{figure*}

\subsection{Cyclic Enhancement-Supervision Joint Optimization}

The key innovation of our approach is the \textbf{cyclic enhancement-supervision mechanism}, which dynamically integrates image enhancement with 3D reconstruction in a mutually reinforcing loop.
This mechanism targets a core challenge in low-light 3D reconstruction: a vicious cycle. Poor image quality hinders accurate geometry learning. In turn, inaccurate geometry limits the effectiveness of image enhancement.
Our cyclic mechanism breaks this deadlock by allowing the ground truth to evolve progressively during training, as the enhancement network and 3D model co-adapt.

High-quality 3D reconstruction requires well-enhanced images, but effective image enhancement relies on accurate 3D geometry. To solve this, we introduce a novel \textbf{evolving supervision strategy} where the ground truth $GT^{(t)}$ is updated at each training iteration based on the output of the enhancement network.

\subsubsection{Initialization and Cyclic Update Mechanism}

At the start of training, the raw low-light images are used as the initial ground truth:
\begin{equation}
GT^{(0)} = I_{\text{low}}.
\end{equation}
This serves as the starting point for the 3D Gaussian Splatting model. At each iteration $t$, the enhancement network $\mathcal{R}$ (Retinex) generates an enhanced image:
\begin{equation}
I_{\text{enh}}^{(t)} = \mathcal{R}(I_{\text{low}}).
\end{equation}
The ground truth is then updated by blending the previous ground truth and the current enhanced output:
\begin{equation}
GT^{(t)} = (1 - \alpha^{(t)}) \cdot GT^{(t-1)} + \alpha^{(t)} \cdot I_{\text{enh}}^{(t)},
\end{equation}
where the blending factor $\alpha^{(t)}$ increases linearly to ensure smooth transition:
\begin{equation}
\tag{4}\label{eq:alpha_linear}
\alpha^{(t)} = \min\left(1.0, \frac{t}{T_{\text{transition}}}\right), 
\quad T_{\text{transition}} = 8{,}000.
\end{equation}
the enhancement network typically converges around 5k–10k iterations independent of total training steps, a fixed threshold of 8,000 triggers the raw→enhanced supervision handover right after convergence—avoiding premature switches in short runs and unnecessary delays in long runs—yielding robust, efficient performance across scenes and budgets.

This multi-stage update ensures:

\textbf{Early stage:} Supervision relies primarily on raw input to avoid unstable enhancement artifacts.
\textbf{Middle stage:} A balanced combination of raw and enhanced images stabilizes joint optimization.
\textbf{Late stage:} Enhanced and thermally corrected images dominate, improving final radiometric and geometric quality.

\subsubsection{Thermal \& Geometric Losses with Adaptive Weighting}

We define the total loss for the 3D Gaussian Splatting model as a combination of terms that balance image quality, structural consistency, and thermal fidelity. 


To ensure radiometric coherence between the enhanced images and 3D reconstructions, 
we introduce a \textbf{thermal consistency loss} that preserves both the enhancement 
quality and the inherent thermal characteristics of the original imagery. Unlike 
conventional reconstruction losses that only measure pixel-wise similarity, our 
thermal consistency loss enforces a dual constraint that balances enhancement 
fidelity with thermal integrity:

\begin{equation}
\tag{6}\label{eq:thermal_loss}
\begin{aligned}
\mathcal{L}_{\text{therm}} 
&= \gamma \left\| \Phi_{\text{rgb}}(I_{\text{enh}}) - \Phi_{\text{rgb}}(I_{\text{rendered}}) \right\|_1 \\
&\quad + (1-\gamma) \left\| \Phi_{\text{rgb}\leftrightarrow\text{therm}}(I_{\text{enh}}) - \Phi_{\text{therm}}(I_{\text{low}}^{\text{therm}}) \right\|_1,
\end{aligned}
\end{equation}
Here $I_{\text{low}}^{\text{therm}}$ denotes the original thermal infrared image (distinct from the low-light RGB $I_{\text{low}}$). 
We use lightweight cross-modal alignment to place both modalities in a shared metric space before the $\ell_1$:
$\Phi_{\text{rgb}}$ maps RGB to per-image luminance with min--max normalization,
$\Phi_{\text{therm}}$ min--max normalizes the thermal image to $[0,1]$ (or temperature range if calibrated),
and $\Phi_{\text{rgb}\leftrightarrow\text{therm}}$ is the same luminance+normalization applied to $I_{\text{enh}}$ so it matches the thermal range.
This makes the two terms dimensionally consistent without changing network outputs. 
We set $\gamma=0.1$.

The first term $\gamma \left\| \Phi_{\text{rgb}}(I_{\text{enh}}) - \Phi_{\text{rgb}}(I_{\text{rendered}}) \right\|_1$ enforces 
\textbf{enhancement-reconstruction consistency}, ensuring that the 3D Gaussian 
Splatting renderer produces outputs that align with the enhanced images. This 
cross-modal consistency is critical for preventing the "enhancement–geometry 
deadlock," where improved image quality leads to geometric artifacts, or vice versa. 
By maintaining this alignment, we ensure that geometric learning benefits from 
enhanced visual features while remaining stable.

The second term provides \textbf{thermal preservation regularization}, preventing the enhancement 
network from over-adjusting the original thermal distribution. This constraint is 
particularly important for thermal imaging, where radiometric values carry physical 
meaning (e.g., temperature information). By penalizing excessive deviation from the 
raw input, we preserve the semantic content of thermal data while allowing necessary 
brightness and contrast improvements.

To ensure stable convergence and balanced optimization across multiple objectives, 
we formulate the \textbf{total training objective} that integrates three major 
components with \textbf{adaptive weighting}:

\begin{equation}
\tag{7}\label{eq:total_loss}
\begin{aligned}
\mathcal{L}_{\text{total}}^{(t)} 
&= \lambda_{\text{enh}}^{(t)} \,\mathcal{L}_{\text{enh}}(I_{\text{enh}}, I_{\text{low}}) \\
&\quad + \lambda_{\text{gs}}^{(t)} \,\mathcal{L}_{\text{gs}}(I_{\text{rendered}}, GT^{(t)}) \\
&\quad + \lambda_{\text{therm}}^{(t)} \,\mathcal{L}_{\text{therm}},
\end{aligned}
\end{equation}

where $GT^{(t)}$ evolves according to Eq.~(3) with the linear blending schedule 
in Eq.~(4), and the adaptive weights $\{\lambda_{\text{enh}}^{(t)}, \lambda_{\text{gs}}^{(t)}, 
\lambda_{\text{therm}}^{(t)}\}$ are dynamically adjusted throughout training.

This scheduling allows us to shift focus from enhancing images early in training to refining geometry and thermal consistency later in the process.

\subsubsection{Convergence and Benefits of the Cyclic Mechanism}


The proposed cyclic training paradigm guarantees convergence by leveraging three core properties. 
Monotonicity: As the reliability weight $\alpha^{(t)}$ increases, the supervision progressively emphasizes higher-quality ground truth, ensuring a monotonic improvement of reconstruction fidelity. 
Stability: The gradual update strategy mitigates abrupt optimization shifts, thereby maintaining training stability. 
Adaptivity: The supervision dynamically aligns with the model’s evolving prediction capability, facilitating self-consistent refinement.

Overall, this cyclic mechanism fosters mutual enhancement between image restoration and 3D geometry estimation, yielding reconstructions that are not only geometrically precise but also thermally coherent, even in complex low-illumination scenarios.

\subsection{Retinex-Driven Joint Optimization with 3D Gaussian Splatting}

To achieve illumination-aware 3D reconstruction under extremely low-light conditions, we formulate DTGS as a unified optimization framework that couples Retinex-based enhancement, thermal supervision, and 3D Gaussian Splatting (3DGS). Unlike conventional pipelines that apply enhancement as an independent preprocessing step, DTGS embeds Retinex decomposition directly inside the reconstruction loop, allowing illumination correction and geometric modeling to evolve jointly under thermal constraints.

\subsubsection{Retinex formulation within 3DGS.}
Given a low-light input image \(I_{\text{low}}\in\mathbb{R}^{H\times W\times 3}\), 
DTGS decomposes it into reflectance \(R\) and illumination \(L\) following the Retinex principle:
\[
I_{\text{low}} = R \odot L,\qquad I_{\text{enh}} = R \odot L^{\prime}.
\]
“$L'$ denotes the illumination after correction.

To ensure smooth transition during training, we employ a \textbf{four-stage piecewise linear scheduling} mechanism for loss weights:

\begin{equation}
\tag{8}\label{eq:lambda_sched}
\lambda_k^{(t)} = 
\begin{cases}
\lambda_k^{(0)}, & \text{if } t/T < 0.2 \\
\lambda_k^{(0)} + (\lambda_k^{(\text{final})} - \lambda_k^{(0)}) \cdot \frac{t/T - 0.2}{0.2}, & \text{if } 0.2 \le t/T < 0.4 \\
\lambda_k^{(\text{final})}, & \text{if } 0.4 \le t/T < 0.7 \\
\text{fine-tuned}, & \text{if } t/T \ge 0.7
\end{cases}
\end{equation}

where $T$ is the total number of iterations, and $k \in \{\text{enh}, \text{gs}, \text{therm}\}$. 
This four-stage design allows the model to: (1) focus on enhancement initialization, 
(2) smoothly balance all objectives, (3) stabilize 3D reconstruction, and 
(4) fine-tune radiometric consistency. The weights satisfy the normalization constraint:
\begin{equation}
\tag{9}\label{eq:weight_constraint}
\lambda_{\text{enh}}^{(t)} + \lambda_{\text{gs}}^{(t)} + \lambda_{\text{therm}}^{(t)} = 1,
\quad \lambda_{\text{gs}}^{(t)} \ge 0.1.
\end{equation}
and the enhanced image is reconstructed as $I_{\text{enh}} = R \odot L'$.  
This decomposition explicitly separates exposure-variant illumination from exposure-invariant reflectance, producing physically interpretable enhancement that aligns with 3DGS appearance parameters (color and opacity).
We initialize the task-level weight of the thermal branch at $\lambda_{\text{therm}}^{(0)} = 0.1$ and schedule it via Eq.~(9) under the normalization constraint in Eq.~(10).

When the thermal branch is enabled, task weights are always renormalized to satisfy Eq.~(9).
For the implementation detail “Retinex/3DGS set to $0.1/0.9$ with an additional thermal-specific $0.2$”, we use normalized weights $(\lambda_{\text{enh}},\lambda_{\text{gs}},\lambda_{\text{therm}}) = \frac{(0.1,0.9,0.2)}{0.1+0.9+0.2} \approx (0.083,\,0.750,\,0.167)$ to ensure $\lambda_{\text{enh}}+\lambda_{\text{gs}}+\lambda_{\text{therm}}=1$.


\subsubsection{Coupling with 3D Gaussian Splatting.}
The enhanced image $I_{\text{enh}}$ serves as supervision for the 3DGS renderer, which represents the scene as a set of $N$ Gaussians $\mathcal{G}=\{(\mu_i,\Sigma_i,\alpha_i,c_i)\}_{i=1}^N$.  
Rendering is performed via front-to-back alpha compositing:
\begin{equation}
\tag{10}
C(r) = \sum_{i=1}^{N(r)} c_i \alpha_i \prod_{j<i}(1-\alpha_j),
\end{equation}
where $C(r)$ is the color of ray $r$ intersecting the Gaussian set.  
Loss gradients from 3D reconstruction propagate through the Retinex branch, forcing it to favor decompositions that yield geometry-consistent enhancement and stable cross-view illumination.  
This closed-loop coupling avoids the ``enhancement–geometry deadlock'' common in two-stage systems.

\section{RGBT-LOW Dataset}

\begin{figure*}
    \centering
    \includegraphics[width=1\linewidth]{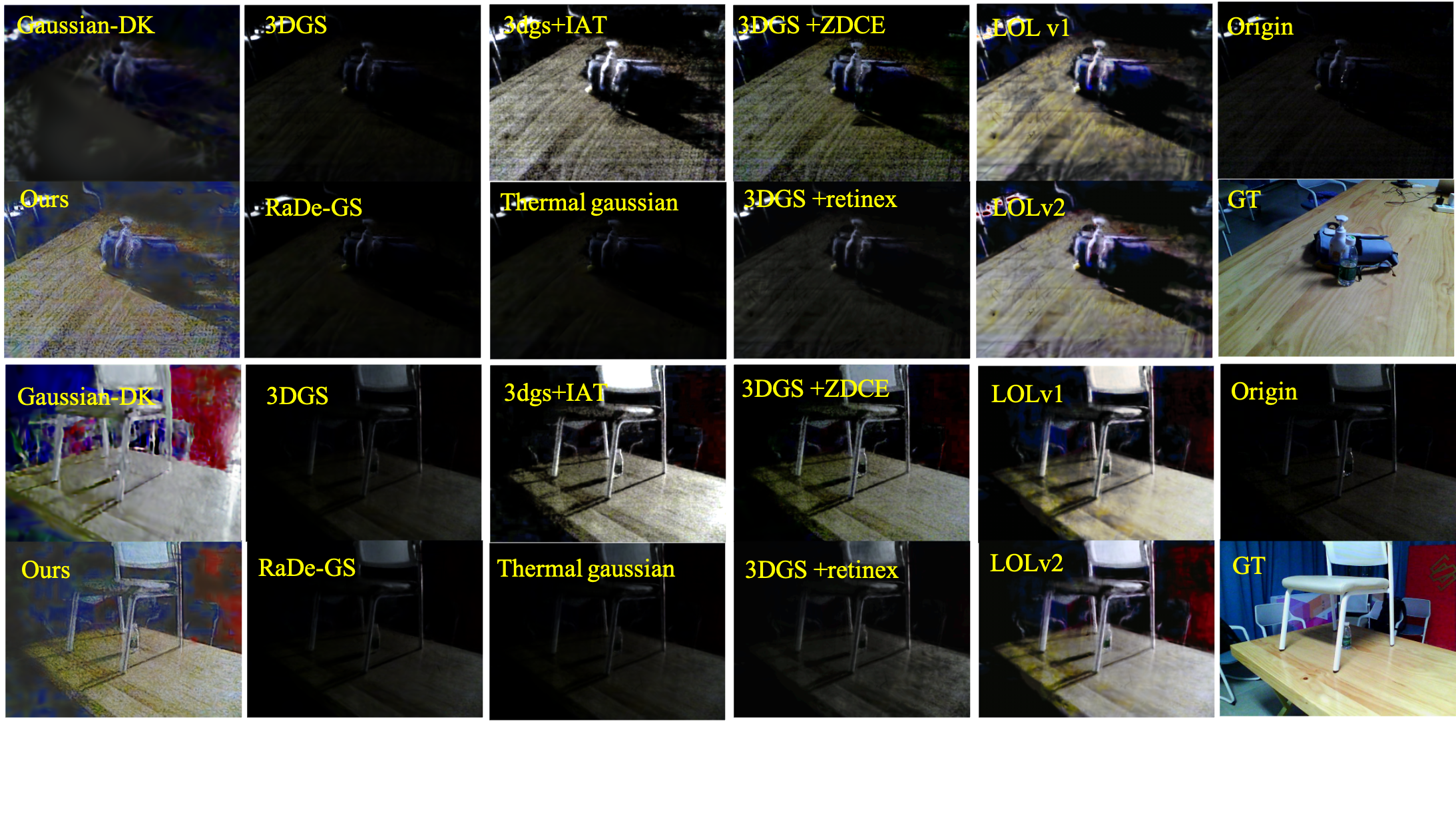}
    \caption{
    Qualitative comparison on the RGBT-LOW dataset. 
    Our method produces more consistent color restoration and structural fidelity across views. 
    Compared with other enhancement-based approaches, DTGS effectively preserves object integrity and avoids enhancement tearing, geometric distortion, or color shifts commonly seen in partially enhanced regions.include gaussian-DK~\cite{ye2024gaussian},3dgs,retinexformer, RaDe-GS~\cite{zhang2024rade},Thermal Gaussian
    }
    \label{fig:placeholder2}
\end{figure*}
To rigorously assess 3D reconstruction in low-light thermal conditions, we introduce the \textbf{RGBT-LOW} dataset. Existing benchmarks—such as those used in Thermal Gaussian—capture RGB images in well-lit environments, failing to represent severe real-world degradation. Datasets collected in dim settings often lack an additional modality like LOM~\cite{cui2024alethnerfilluminationadaptivenerf}, limiting their utility. RGBT-LOW addresses this gap by providing RGB scenes with minimal visible detail, forcing reconstruction methods to depend primarily on thermal cues.

The \textbf{RGBT-LOW} dataset consists of 20 real-world scenes, with a total of 6000 images, including various objects such as computer chairs, bottles, bags, and books. This amounts to a total of 75$\times$2$\times$20$\times$2 synchronized RGB (dark light corresponding to bright light GT) and thermal image pairs. For each viewpoint, we generate low-light and normal-light images, while keeping other camera settings unchanged. We capture multi-view images by moving and rotating the camera mount. The image acquisition resolution is 680$\times$480. Each scene contains approximately 75$\times$2$\times$2 images captured using \textbf{FLIR} cameras. Specifically, all images were captured using the \textbf{FLIR E6} camera model, which can simultaneously capture both RGB and thermal images.

To ensure fair comparisons, we use indoor lighting supplementation to collect the GT data and employ the same method for training, segmentation, and testing. We provide the original images captured by the thermal camera, along with RGB images, thermal images, MSX images, and camera pose data. Our dataset ensures consistency in thermal measurements across different viewpoints and covers a wide range of scenes. For more detailed information about our dataset, please refer to the supplementary materials.

\section{Experiment}
We evaluate all methods on the proposed RGBT-LOW dataset under consistent settings.

\begin{table*}[t]
\centering
\caption{
Quantitative comparison of image quality and perceptual metrics, including SSIM,PSNR, and LPIPS,across four object classes and the overall mean. Section (A) reports 3D/thermal-only baselines, (B) presents pipelines combining low-light image enhancement with 3D Gaussian Splatting (3DGS), and (C) shows our proposed \textbf{DTGS } method. For fair comparison with recent low-light enhancement approaches, we additionally evaluate \textbf{Retinexformer} models like $ LOLv1$  and$  LOLv2real$  released in 2024–2025 (ECCV 2024 version and updates announced for the NTIRE 2025 Low-Light Image Enhancement Challenge.).The mean in this table is the average of multiple datasets, not just the datasets listed in the four columns below.\textit{A detailed analysis of the selected baselines for comparison is provided in the Supplementary Material.}
}
\setlength{\tabcolsep}{3pt}
\resizebox{\textwidth}{!}{
\begin{tabular}{lcccccccccccccccc}
\toprule
\multirow{2}{*}{Method} &
\multicolumn{3}{c}{Books} &
\multicolumn{3}{c}{Bottle} & 
\multicolumn{3}{c}{Chair} &
\multicolumn{3}{c}{Computer} &
\multicolumn{3}{c}{\textbf{mean}} \\
\cmidrule(lr){2-4}\cmidrule(lr){5-7}\cmidrule(lr){8-10}\cmidrule(lr){11-13}\cmidrule(lr){14-16}
& SSIM \(\uparrow\) & PSNR \(\uparrow\) & LPIPS \(\downarrow\) & SSIM \(\uparrow\) & PSNR \(\uparrow\) & LPIPS \(\downarrow\) & SSIM \(\uparrow\) & PSNR \(\uparrow\) & LPIPS \(\downarrow\) & SSIM \(\uparrow\) & PSNR \(\uparrow\) & LPIPS \(\downarrow\) &
SSIM \(\uparrow\) & PSNR \(\uparrow\) & LPIPS \(\downarrow\) \\

\cmidrule(lr){1-1} \cmidrule(lr){2-4}\cmidrule(lr){5-7}\cmidrule(lr){8-10}\cmidrule(lr){11-13}\cmidrule(lr){14-16}
\multicolumn{16}{l}{\emph{(A) 3D / Thermal-only baselines}} \\
3DGS & 0.2470 & 7.61 & 0.6844 & 0.0914 & 6.91 & 0.6793 & 0.1725 & 7.12 & 0.6802 & 0.1386 & 7.04 & 0.6837 & 0.1516 & 7.09 & 0.6865 \\
RaDe-GS               & 0.2518 & 7.63 & 0.6891 & 0.0893 & 6.89 & 0.6813 & 0.1736 & 7.13 & 0.6802 & 0.1402 & 7.05 & 0.6846 & 0.1526 & 7.09 & 0.6882 \\
Thermal Gaussian       & 0.2518 & 7.63 & 0.6891 & 0.1212 & 7.12 & 0.6811 & 0.1744 & 7.22 & 0.6721 & 0.0516 & 6.36 & 0.6932 & 0.1416 & 7.04 & 0.6877 \\
Gaussian-DK           & 0.1886 & 7.63 & 0.6869 & 0.0755 & 6.96 & 0.6993 & 0.1271 & 7.15 & 0.6825 & 0.1043 & 7.03 & 0.7063 & 0.1134 & 7.09 & 0.7006 \\
\midrule
\addlinespace
\multicolumn{16}{l}{\emph{(B) Pre-enhance $\rightarrow$ 3DGS}} \\
3DGS + IAT            & 0.2875 & 9.13 & 0.6623 & 0.1669 & 8.45 & 0.6875 & 0.2089 & 8.19 & 0.6790 & 0.2298 & 8.73 & 0.6791 & 0.2139 & 8.58 & 0.6811 \\
3DGS + Zero-DCE       & 0.3010 & 10.18 & 0.6573 & 0.1381 & 7.85 & 0.6786 & 0.2045 & 8.74 & 0.6688 & 0.2120 & 8.70 & 0.6628 & 0.2034 & 8.69 & 0.6697 \\
Zero-DCE + 3DGS       & 0.2920 & 9.30 & 0.6532 & 0.1381 & 7.92 & 0.6786 & 0.2055 & 8.92 & 0.6642 & 0.2298 & 8.92 & 0.6791 & 0.2053 & 8.68 & 0.6674 \\
3DGS + Retinex        & 0.2790 & 8.41 & 0.6844 & 0.1631 & 7.40 & 0.6866 & 0.2299 & 7.71 & 0.6846 & 0.1978 & 7.59 & 0.6905 & 0.2076 & 7.67 & 0.6914 \\
3DGS +LOL\_v1               & 0.3444 & 9.32 & 0.6636 & 0.2101 & 8.23 & 0.6865 & 0.2668 & 8.42 & 0.6792 & 0.2869 & 8.74 & 0.6845 & 0.2776 & 8.68 & 0.6789 \\
3DGS +LOL\_v2\_real          & 0.3498 & 9.66 & 0.6697 & 0.2552 & 8.89 & 0.6887 & 0.2774 & 8.73 & 0.6752 & 0.3061 & 9.26 & 0.6857 & 0.2971 & 9.13 & 0.6798 \\
\midrule
\addlinespace
\multicolumn{16}{l}{\emph{(C) Ours}} \\
\textbf{DTGS}  & \textbf{0.3251} & \textbf{11.07} & \textbf{0.6344} &
\textbf{0.4842} & \textbf{11.17} & \textbf{0.6771} &
\textbf{0.2689} & \textbf{10.64} & \textbf{0.6399} &
\textbf{0.3445} & \textbf{11.70} & \textbf{0.6490} &
\textbf{0.3520} & \textbf{11.17} & \textbf{0.6520} \\
\bottomrule
\end{tabular}}
\footnotesize{}
\label{tab:quantitative_results}
\end{table*}

\subsection{Implementation Details}

All experiments are conducted on NVIDIA RTX 3090 GPUs. The model is trained for 30,000 iterations with an image resolution of $680 \times 480$. We use the Adam optimizer with an initial learning rate of $1 \times 10^{-3}$ and apply cosine decay every 5000 iterations. The 3D Gaussian positions are initialized using COLMAP for camera pose estimation. In our joint training framework, the loss weights for RetinexNet and 3D Gaussian Structures (3DGS) are set to 0.1 and 0.9, respectively, along with a thermal-specific loss weight of 0.2 to address the unique characteristics of thermal images.

A cyclic ground truth (GT) update mechanism is employed, with a transition period of 8000 iterations. The overall loss function is a weighted combination of several components: L1 loss (0.7), SSIM loss (0.2), Corner loss (0.1), and consistency loss (0.1). In addition, adaptive weight scheduling is applied to RetinexNet. During early training (0--30\%), the weight is set to 0.3, which decreases to 0.2 in the mid-training phase (30--70\%) and further reduces to 0.1 in the later stages (70--100\%), ensuring a smooth transition from image enhancement to geometric refinement.

\subsection{Comprehensive Evaluation}
We conduct a comprehensive evaluation of novel view synthesis under low-light thermal conditions using the RGBT-LOW dataset. Initially, we evaluate the original 3DGS and RaDe-GS Thermal Gaussian on their ability to synthesize new views. Following this, to assess the effectiveness of our proposed end-to-end method in enhancing dark-light scenarios and preserving color consistency, we incorporate Zero-DCE and IAT for the original 3DGS and Gaussian-DK, which is another end-to-end network designed for enhancing dark-light thermal data.All enhancement models (IAT and Zero-DCE) are applied with their default pretrained weights and configurations.

\subsubsection{Experimental Analysis}
\label{sec:exp_analysis}

We conduct a comprehensive evaluation of the proposed \textbf{DTGS} framework under low-light conditions, comparing it against a range of state-of-the-art 3D Gaussian Splatting (3DGS) and low-light enhancement methods. Quantitative results are summarized in \textbf{Table~\ref{tab:quantitative_results}}.

\paragraph{(A) 3D / Thermal-only Baselines.}
Traditional Gaussian-based rendering methods, including \textbf{3DGS}, \textbf{RaDe-GS}, \textbf{Thermal Gaussian}, and \textbf{Gaussian-DK} (``Gaussian in the Dark''), demonstrate highly limited capability in extremely dark scenes, with average SSIM around \textbf{0.15}, PSNR around \textbf{7~dB}, and LPIPS exceeding \textbf{0.68}. 
Notably, \textbf{Gaussian-DK} exhibits severe color inconsistency, geometric instability, and blurred rendering when applied to end-to-end low-light enhancement, indicating that the absence of an explicit illumination adaptation mechanism prevents these models from maintaining consistent geometry and texture fidelity under dark illumination. 
Even when thermal cues are available, conventional Gaussian rendering fails to preserve both \emph{textural detail} and \emph{lighting consistency} simultaneously.

\paragraph{(B) Pre-enhancement $\rightarrow$ 3DGS.}
To mitigate visibility degradation in low-light scenarios, we further evaluate combinations of representative enhancement models (\textbf{IAT}, \textbf{Zero-DCE}, \textbf{Retinex}, and \textbf{LOL} variants) as preprocessing modules before 3DGS reconstruction. 
Hybrid pipelines such as \textbf{3DGS+IAT} and \textbf{3DGS+Zero-DCE} show moderate gains in certain scenes (e.g., \emph{books}, \emph{computer}), with PSNR improving to \textbf{8.5--10~dB} and SSIM rising to \textbf{0.20--0.30}. 
However, their performance varies substantially across object categories. 
In complex or severely dark environments, they often cause \emph{texture misalignment}, \emph{geometric drift}, and \emph{blurry point clouds}. 
For instance, both IAT and Zero-DCE fail to recover the red background in the \emph{chair} scene, while \textbf{Gaussian-DK} produces local overexposure. 
Although recent transformer-based \textbf{Retinex} models (e.g., \textbf{LOL\_v1}, \textbf{LOL\_v2\_real}) achieve improved brightness restoration and contrast balance (PSNR $\approx$ 8.7--9.1~dB, SSIM $\approx$ 0.27--0.30), they remain prone to \emph{noise amplification} and \emph{inconsistent color enhancement}, revealing unstable generalization in complex illumination.

\paragraph{(C) Ours: DTGS}
\noindent In contrast, our proposed DTGS achieves the best results across all datasets and metrics, with an average SSIM = 0.3520, PSNR = 11.17 dB, and LPIPS = 0.6520.
Notably, relative to the strongest competing pipelines, DTGS improves the average SSIM by \textbf{+18.5\%}, the average PSNR by \textbf{+22.3\%}, and reduces LPIPS by \textbf{-2.3\%}.
On challenging scenes such as \emph{books} and \emph{bottle}, DTGS yields PSNR gains of +0.89 dB and +2.28 dB; for \emph{chair} and \emph{computer}, the gains are +1.72 dB and +2.44 dB, respectively.
These consistent improvements highlight DTGS’s capability to reconstruct fine geometry while maintaining thermal–visual coherence under extremely low-light conditions.

This remarkable performance stems from DTGS’s cyclic enhancement–supervision mechanism, which enforces bidirectional consistency between the visual and thermal domains: (1) the enhancement branch leverages thermal cues to guide illumination restoration and noise suppression in the visual branch; (2) the supervision branch projects the enhanced visual features back to the thermal space to regularize geometry and structure; and (3) cyclic consistency losses align both domains at low-level (illumination/edges) and high-frequency (textures) scales. This dual-domain supervision enables effective recovery of low-level illumination cues and high-frequency detail while preserving heat-aware structural integrity.

\paragraph{Qualitative Evaluation.}
As illustrated in \textbf{Figure~\ref{fig:placeholder2}}, baseline methods such as \textbf{3DGS} and \textbf{RaDe-GS} produce blurred reconstructions with illumination artifacts, while enhancement-based variants such as \textbf{IAT} and \textbf{Retinexformer} yield partially recovered structures yet suffer from \emph{uneven textures} and \emph{color inconsistency}. 
In contrast, \textbf{DTGS} delivers \emph{sharper edges}, \emph{uniform brightness}, and \emph{high-fidelity detail reconstruction}—for instance, clearly restoring the background panel and bottle contours. 
These results demonstrate that DTGS sets a new performance benchmark for low-light multimodal 3D Gaussian rendering, ensuring both \textbf{geometric accuracy} and \textbf{perceptual consistency} across diverse lighting conditions.

\subsection{Ablation Study}
To investigate the contribution of each module to the low-light thermal 3D reconstruction task, we perform a systematic ablation study on the RGBT-LOW dataset. The results are summarized in Table~\ref{tab:ablation-vertical}. Notably, \textbf{Thermal Gaussian} performs the worst across all metrics (average SSIM of only 0.1416, PSNR of 7.04 dB), showing that relying solely on thermal modality in extremely dark scenarios leads to significant information loss. Introducing cyclic supervision (\textbf{Thermal w/o Cyclic}) results in improvements in stability and light consistency, with a 3 dB increase in PSNR on average, validating the importance of cyclic GT updates for joint optimization of geometry and thermal features.

\begin{table}[t]
\centering
\caption{Ablation Study}
\label{tab:ablation-vertical}
\renewcommand{\arraystretch}{1.12}
\setlength{\tabcolsep}{4pt}
\resizebox{\linewidth}{!}{%
\begin{tabular}{llcccc}
\toprule
\multirow{2}{*}{Class} & \multirow{2}{*}{Metric} & \multicolumn{4}{c}{Method} \\
\cmidrule(lr){3-6}
 & & 3DGS + Retinex & ours w/o Cyclic & Ours & Thermal Gaussian \\
\midrule
bag & SSIM$\uparrow$      & 0.2274& \second{0.2431} & \best{0.3368} & 0.0716 \\
    & PSNR$\uparrow$      & 7.67& \second{10.37}  & \best{11.26}  & 6.65 \\
    & LPIPS$\downarrow$   & 0.6897& \second{0.6714} & \best{0.6585} & 0.7281 \\
\midrule
books & SSIM$\uparrow$    & 0.2790               & 0.2104          & \best{0.3251} & \second{0.2518} \\
      & PSNR$\uparrow$    & 8.41                 & 8.90            & \best{11.07}  & \second{7.63} \\
      & LPIPS$\downarrow$ & 0.6844               & 0.6651          & \best{0.6344} & \second{0.6891} \\
\midrule
bottle & SSIM$\uparrow$   & 0.1631               & 0.3207          & \best{0.4842} & \second{0.0893} \\
       & PSNR$\uparrow$   & 7.40                 & 10.95           & \best{11.17}  & \second{6.89} \\
       & LPIPS$\downarrow$& 0.6866               & \best{0.6756}   & \second{0.6771} & 0.6813 \\
\midrule
chair & SSIM$\uparrow$    & 0.2299               & 0.1937          & \best{0.2689} & \second{0.1736} \\
      & PSNR$\uparrow$    & 7.71                 & 9.34            & \best{10.64}  & \second{7.13} \\
      & LPIPS$\downarrow$ & 0.6846               & 0.6617          & \best{0.6399} & \second{0.6802} \\
\midrule
computer & SSIM$\uparrow$ & 0.1978               & \second{0.2283} & \best{0.3445} & 0.1402 \\
         & PSNR$\uparrow$ & 7.59                 & \second{10.00}  & \best{11.70}  & 7.05 \\
         & LPIPS$\downarrow$ & 0.6905             & \second{0.6783} & \best{0.6490} & 0.6846 \\
\midrule
mean & SSIM$\uparrow$     & 0.2076               & 0.2392          & \best{0.3520} & \second{0.1453} \\
     & PSNR$\uparrow$     & 7.67                 & 9.91            & \best{11.17}  & \second{7.07} \\
     & LPIPS$\downarrow$  & 0.6914               & 0.6704          & \best{0.6520} & \second{0.6927} \\
\bottomrule
\end{tabular}
}
\end{table}

\begin{figure}
    \centering
    \includegraphics[width=1\linewidth]{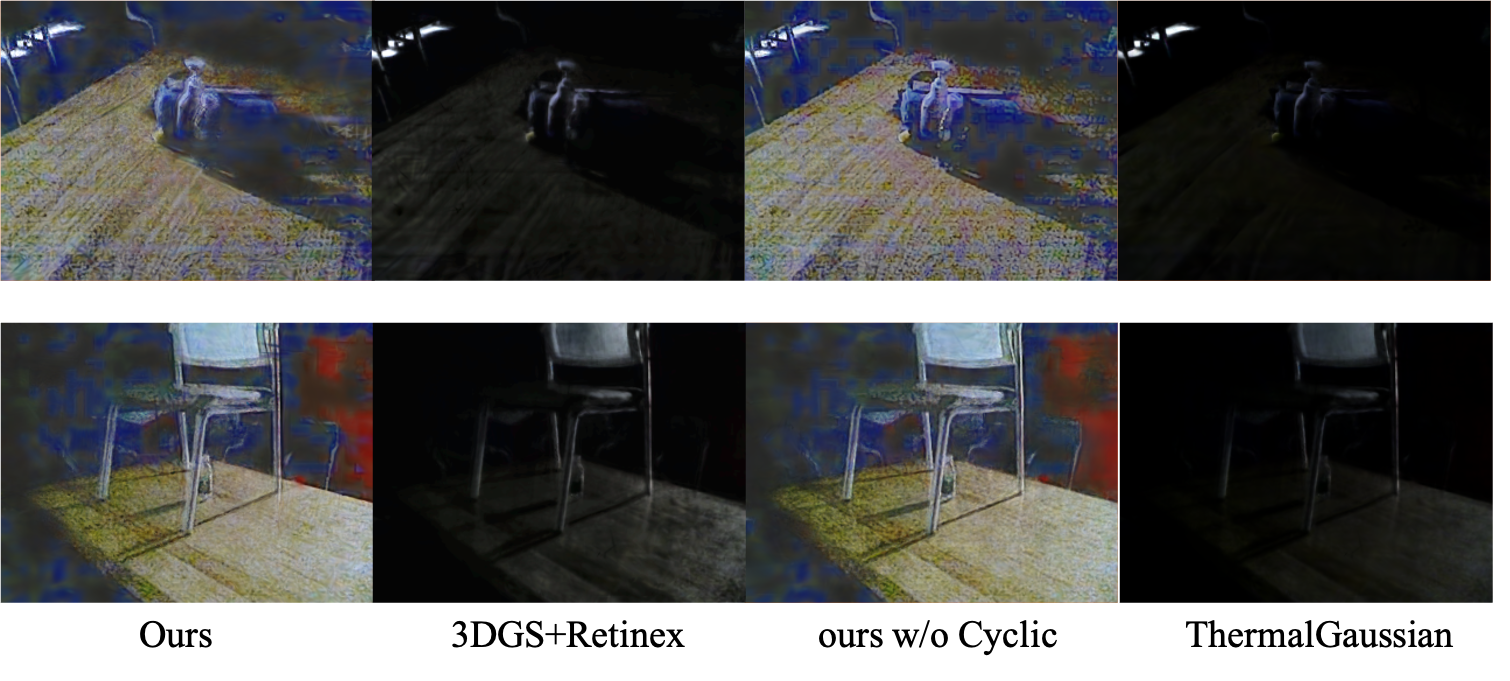}
    \caption{Qualitative comparison of ablation variants.}
    \label{fig:ablation_qualitative}
\end{figure}

Further, our method without the cyclic supervision (\textbf{Ours w/o Cyclic}) already demonstrates significant improvements over all baseline variants, confirming the effectiveness of the proposed thermal-aware cyclic enhancement mechanism. Compared with Thermal Gaussian and the pre-enhancement baseline (\textbf{3DGS + Retinex}), our DTGS model achieves the best performance, with an average SSIM of 0.3520 and PSNR of 11.17 dB. These results indicate that the integration of thermal guidance with cyclic ground-truth updates effectively mitigates degradation, enhances illumination stability, and maintains geometric consistency under extreme low-light conditions. Overall, the proposed framework establishes a robust foundation for illumination-aware 3D reconstruction.

\section{Conclusion}

We propose \textbf{DTGS}, the first thermal-supervised 3D Gaussian Splatting framework for extreme low light. DTGS unifies 3DGS reconstruction with a Retinex-based enhancer in a joint optimization, stabilized by a thermal branch that injects illumination-invariant structure to guide both reflectance recovery and geometry. A \emph{Cyclic Enhancement–Supervision} scheme refreshes supervision during training so the two modules co-evolve.

{
    \small
    \bibliographystyle{ieeenat_fullname}
    \bibliography{main}

@String(BMVC= {Brit. Mach. Vis. Conf.})

@String(AAAI = {AAAI})

@String(BMVC  =	{BMVC})

@inproceedings{connor2002novel,
  title={Novel View Specification and Synthesis.},
  author={Connor, Keith Richard and Reid, Ian},
  booktitle={BMVC},
  pages={1--10},
  year={2002}
}

@article{habtegebrial2018fast,
  title={Fast view synthesis with deep stereo vision},
  author={Habtegebrial, Tewodros and Varanasi, Kiran and Bailer, Christian and Stricker, Didier},
  journal={arXiv preprint arXiv:1804.09690},
  year={2018}
}

@article{li2207compnvs,
  title   = {CompNVS: Novel View Synthesis with Scene Completion},
  author  = {Li, Zuoyue and Fan, Tianxing and Li, Zhenqiang and Cui, Zhaopeng and Sato, Yoichi},
  journal = {arXiv preprint arXiv:2207.11467},
  year    = {2022},
}

@inproceedings{elata2025novel,
  title={Novel view synthesis with pixel-space diffusion models},
  author={Elata, Noam and Kawar, Bahjat and Ostrovsky-Berman, Yaron and Farber, Miriam and Sokolovsky, Ron},
  booktitle={Proceedings of the Computer Vision and Pattern Recognition Conference},
  pages={26756--26766},
  year={2025}
}

@article{mildenhall2021nerf,
  title={Nerf: Representing scenes as neural radiance fields for view synthesis},
  author={Mildenhall, Ben and Srinivasan, Pratul P and Tancik, Matthew and Barron, Jonathan T and Ramamoorthi, Ravi and Ng, Ren},
  journal={Communications of the ACM},
  volume={65},
  number={1},
  pages={99--106},
  year={2021},
  publisher={ACM New York, NY, USA}
}

@article{kerbl20233d,
  title={3D Gaussian splatting for real-time radiance field rendering.},
  author={Kerbl, Bernhard and Kopanas, Georgios and Leimk{\"u}hler, Thomas and Drettakis, George},
  journal={ACM Trans. Graph.},
  volume={42},
  number={4},
  pages={139--1},
  year={2023}
}

@inproceedings{mildenhall2022nerf,
  title={Nerf in the dark: High dynamic range view synthesis from noisy raw images},
  author={Mildenhall, Ben and Hedman, Peter and Martin-Brualla, Ricardo and Srinivasan, Pratul P and Barron, Jonathan T},
  booktitle={Proceedings of the IEEE/CVF conference on computer vision and pattern recognition},
  pages={16190--16199},
  year={2022}
}

@inproceedings{wang2023lighting,
  title={Lighting up nerf via unsupervised decomposition and enhancement},
  author={Wang, Haoyuan and Xu, Xiaogang and Xu, Ke and Lau, Rynson WH},
  booktitle={Proceedings of the IEEE/CVF international conference on computer vision},
  pages={12632--12641},
  year={2023}
}

@inproceedings{guo2020zero,
  title={Zero-reference deep curve estimation for low-light image enhancement},
  author={Guo, Chunle and Li, Chongyi and Guo, Jichang and Loy, Chen Change and Hou, Junhui and Kwong, Sam and Cong, Runmin},
  booktitle={Proceedings of the IEEE/CVF conference on computer vision and pattern recognition},
  pages={1780--1789},
  year={2020}
}

@article{li2021learning,
  title={Learning to enhance low-light image via zero-reference deep curve estimation},
  author={Li, Chongyi and Guo, Chunle and Loy, Chen Change},
  journal={IEEE transactions on pattern analysis and machine intelligence},
  volume={44},
  number={8},
  pages={4225--4238},
  year={2021},
  publisher={IEEE}
}

@article{zhang2021low,
  title={Low-illumination image enhancement in the space environment based on the DC-WGAN algorithm},
  author={Zhang, Minglu and Zhang, Yan and Jiang, Zhihong and Lv, Xiaoling and Guo, Ce},
  journal={Sensors},
  volume={21},
  number={1},
  pages={286},
  year={2021},
  publisher={MDPI}
}

@inproceedings{cai2023retinexformer,
  title={Retinexformer: One-stage retinex-based transformer for low-light image enhancement},
  author={Cai, Yuanhao and Bian, Hao and Lin, Jing and Wang, Haoqian and Timofte, Radu and Zhang, Yulun},
  booktitle={Proceedings of the IEEE/CVF international conference on computer vision},
  pages={12504--12513},
  year={2023}
}

@article{tang2022improved,
  title={An improved algorithm for low-light image enhancement based on RetinexNet},
  author={Tang, Hao and Zhu, Hongyu and Tao, Huanjie and Xie, Chao},
  journal={Applied Sciences},
  volume={12},
  number={14},
  pages={7268},
  year={2022},
  publisher={MDPI}
}

@article{yang2018low,
  title={Low-light image enhancement based on Retinex theory and dual-tree complex wavelet transform},
  author={Yang, Mao-xiang and Tang, Gui-jin and Liu, Xiao-hua and Wang, Li-qian and Cui, Zi-guan and Luo, Su-huai},
  journal={Optoelectronics Letters},
  volume={14},
  number={6},
  pages={470--475},
  year={2018},
  publisher={Springer}
}

@article{hassan2024thermonerf,
  title={Thermonerf: Multimodal neural radiance fields for thermal novel view synthesis},
  author={Hassan, Mariam and Forest, Florent and Fink, Olga and Mielle, Malcolm},
  journal={arXiv preprint arXiv:2403.12154},
  volume={2},
  number={7},
  year={2024}
}

@article{land1971lightness,
  title={Lightness and retinex theory},
  author={Land, Edwin H and McCann, John J},
  journal={Journal of the Optical society of America},
  volume={61},
  number={1},
  pages={1--11},
  year={1971},
  publisher={Optical Society of America}
}

@article{lu2024thermalgaussian,
  title={Thermalgaussian: Thermal 3d gaussian splatting},
  author={Lu, Rongfeng and Chen, Hangyu and Zhu, Zunjie and Qin, Yuhang and Lu, Ming and Zhang, Le and Yan, Chenggang and Xue, Anke},
  journal={arXiv preprint arXiv:2409.07200},
  year={2024}
}

@misc{cui2024alethnerfilluminationadaptivenerf,
      title={Aleth-NeRF: Illumination Adaptive NeRF with Concealing Field Assumption}, 
      author={Ziteng Cui and Lin Gu and Xiao Sun and Xianzheng Ma and Yu Qiao and Tatsuya Harada},
      year={2024},
      eprint={2312.09093},
      archivePrefix={arXiv},
      primaryClass={cs.CV},
      url={https://arxiv.org/abs/2312.09093}, 
}

@article{wei2018deep,
  title={Deep retinex decomposition for low-light enhancement},
  author={Wei, Chen and Wang, Wenjing and Yang, Wenhan and Liu, Jiaying},
  journal={arXiv preprint arXiv:1808.04560},
  year={2018}
}

@article{cui2022you,
  title={You only need 90k parameters to adapt light: a light weight transformer for image enhancement and exposure correction},
  author={Cui, Ziteng and Li, Kunchang and Gu, Lin and Su, Shenghan and Gao, Peng and Jiang, Zhengkai and Qiao, Yu and Harada, Tatsuya},
  journal={arXiv preprint arXiv:2205.14871},
  year={2022}
}

@inproceedings{wang2023ultra,
  title={Ultra-high-definition low-light image enhancement: A benchmark and transformer-based method},
  author={Wang, Tao and Zhang, Kaihao and Shen, Tianrun and Luo, Wenhan and Stenger, Bjorn and Lu, Tong},
  booktitle={Proceedings of the AAAI conference on artificial intelligence},
  volume={37},
  number={3},
  pages={2654--2662},
  year={2023}
}

@inproceedings{vidas20133d,
  title={3D thermal mapping of building interiors using an RGB-D and thermal camera},
  author={Vidas, Stephen and Moghadam, Peyman and Bosse, Michael},
  booktitle={2013 IEEE international conference on robotics and automation},
  pages={2311--2318},
  year={2013},
  organization={IEEE}
}

@article{weinmann2014thermal,
  title={Thermal 3D mapping for object detection in dynamic scenes},
  author={Weinmann, Martin and Leitloff, Jens and Hoegner, Ludwig and Jutzi, Boris and Stilla, Uwe and Hinz, Stefan},
  journal={ISPRS Annals of the Photogrammetry, Remote Sensing and Spatial Information Sciences},
  volume={2},
  pages={53--60},
  year={2014},
  publisher={Copernicus GmbH}
}

@inproceedings{yamaguchi2017application,
  title={Application of LSD-SLAM for visualization temperature in wide-area environment},
  author={Yamaguchi, Masahiro and Saito, Hideo and Yachida, Shoji},
  booktitle={International Conference on Computer Vision Theory and Applications},
  volume={5},
  pages={216--223},
  year={2017},
  organization={SciTePress}
}

@article{de20223d,
  title={3D radiometric mapping by means of lidar SLAM and thermal camera data fusion},
  author={De Pazzi, Davide and Pertile, Marco and Chiodini, Sebastiano},
  journal={Sensors},
  volume={22},
  number={21},
  pages={8512},
  year={2022},
  publisher={MDPI}
}

@inproceedings{xu2024leveraging,
  title={Leveraging thermal modality to enhance reconstruction in low-light conditions},
  author={Xu, Jiacong and Liao, Mingqian and Kathirvel, Ram Prabhakar and Patel, Vishal M},
  booktitle={European Conference on Computer Vision},
  pages={321--339},
  year={2024},
  organization={Springer}
}

@inproceedings{chen2024thermal3d,
  title={Thermal3D-GS: Physics-induced 3D Gaussians for thermal infrared novel-view synthesis},
  author={Chen, Qian and Shu, Shihao and Bai, Xiangzhi},
  booktitle={European Conference on Computer Vision},
  pages={253--269},
  year={2024},
  organization={Springer}
}

@inproceedings{poggi2022cross,
  title={Cross-spectral neural radiance fields},
  author={Poggi, Matteo and Ramirez, Pierluigi Zama and Tosi, Fabio and Salti, Samuele and Mattoccia, Stefano and Di Stefano, Luigi},
  booktitle={2022 International Conference on 3D Vision (3DV)},
  pages={606--616},
  year={2022},
  organization={IEEE}
}

@article{sinha2024spectralgaussians,
  title={SpectralGaussians: Semantic, spectral 3D Gaussian splatting for multi-spectral scene representation, visualization and analysis},
  author={Sinha, Saptarshi Neil and Graf, Holger and Weinmann, Michael},
  journal={arXiv preprint arXiv:2408.06975},
  year={2024}
}

@inproceedings{thirgood2025hypergs,
  title={Hypergs: Hyperspectral 3d gaussian splatting},
  author={Thirgood, Christopher and Mendez, Oscar and Ling, Erin and Storey, Jon and Hadfield, Simon},
  booktitle={Proceedings of the Computer Vision and Pattern Recognition Conference},
  pages={5970--5979},
  year={2025}
}

@inproceedings{ye2024gaussian,
  title={Gaussian in the Dark: Real-Time View Synthesis From Inconsistent Dark Images Using Gaussian Splatting},
  author={Ye, Sheng and Dong, Zhen-Hui and Hu, Yubin and Wen, Yu-Hui and Liu, Yong-Jin},
  booktitle={Computer Graphics Forum},
  volume={43},
  number={7},
  pages={e15213},
  year={2024},
  organization={Wiley Online Library}
}

@article{zhang2024rade,
  title={Rade-gs: Rasterizing depth in gaussian splatting},
  author={Zhang, Baowen and Fang, Chuan and Shrestha, Rakesh and Liang, Yixun and Long, Xiaoxiao and Tan, Ping},
  journal={arXiv preprint arXiv:2406.01467},
  year={2024}
}

@inproceedings{zhou2025lita,
  title={LITA-GS: Illumination-Agnostic Novel View Synthesis via Reference-Free 3D Gaussian Splatting and Physical Priors},
  author={Zhou, Han and Dong, Wei and Chen, Jun},
  booktitle={Proceedings of the Computer Vision and Pattern Recognition Conference},
  pages={21580--21589},
  year={2025}
}

@inproceedings{cui2025luminance,
  title={Luminance-GS: Adapting 3D Gaussian Splatting to Challenging Lighting Conditions with View-Adaptive Curve Adjustment},
  author={Cui, Ziteng and Chu, Xuangeng and Harada, Tatsuya},
  booktitle={Proceedings of the Computer Vision and Pattern Recognition Conference},
  pages={26472--26482},
  year={2025}
}
}


\end{document}